\documentclass{article}

    \usepackage[preprint]{neurips_2025}

\usepackage[utf8]{inputenc} 
\usepackage[T1]{fontenc}    
\usepackage{hyperref}       
\usepackage{url}            
\usepackage{booktabs}       
\usepackage{amsfonts}       
\usepackage{nicefrac}       
\usepackage{microtype}      
\usepackage{xcolor}         

\usepackage{enumitem}
\usepackage{comment}
\usepackage{amssymb}

\usepackage{graphicx}
\usepackage{amsmath} 
\usepackage{float}
\usepackage{wrapfig}

\usepackage{multirow}
\usepackage{multicol}
\usepackage{makecell}

\usepackage{pifont}
\usepackage{microtype}
\usepackage{graphicx}
\usepackage{subfigure}
\usepackage{booktabs} 

\usepackage[ruled,vlined]{algorithm2e}
\usepackage{algorithmic}

\usepackage{caption}

\usepackage{amsthm}
\theoremstyle{plain}
\newtheorem{theorem}{Theorem}[section]

\newtheorem{definition}[theorem]{Definition}

\newtheorem{remark}[theorem]{Remark}

\newcommand{\wt}{\widetilde}

\renewcommand{\tilde}{\wt}

\title{
DraftAttention: Fast Video Diffusion via Low-Resolution Attention Guidance
}

\author{
  Xuan Shen$^{1*}$, \ Chenxia Han$^{2*}$, \ Yufa Zhou$^{3}$\thanks{Equal Contribution} \ , \ Yanyue Xie$^1$, \ Yifan Gong$^4$,
  \\
  \textbf{
  Quanyi Wang$^5$, \ Yiwei Wang$^6$, \
  Yanzhi Wang$^1$\thanks{Corresponding Authors} \ , \ Pu Zhao$^{1\dag}$, \ Jiuxiang Gu$^{4\dag}$
  }
  \\
  $^1$Northeastern University, $^2$CUHK, $^3$Duke University, 
  \\
  $^4$Adobe Research, $^5$NUIST, $^6$UCM
  \\
  \texttt{shen.xu@northeastern.edu},  \texttt{cxhan@cse.cuhk.edu.hk}
}

\newif\ifmodify 
\modifytrue 
\ifmodify

\else

\fi

\begin{document}

\maketitle

\begin{abstract}

Diffusion transformer–based video generation models (DiTs) have recently attracted widespread attention for their excellent generation quality.
However, their computational cost remains a major bottleneck—attention alone accounts for over 80\% of total latency, and generating just 8 seconds of 720p video takes tens of minutes—posing serious challenges to practical application and scalability.
To address this, we propose the \texttt{DraftAttention}, a training-free framework for the acceleration of video diffusion transformers with dynamic sparse attention on GPUs.
We apply down-sampling to each feature map across frames in the compressed latent space, enabling a higher-level receptive field over the latent composed of hundreds of thousands of tokens. 
The low-resolution draft attention map, derived from draft query and key, exposes redundancy both spatially within each feature map and temporally across frames.
We reorder the query, key, and value based on the draft attention map to guide the sparse attention computation in full resolution, and subsequently restore their original order after the attention computation.
This reordering enables structured sparsity that aligns with hardware-optimized execution.
Our theoretical analysis demonstrates that the low-resolution draft attention closely approximates the full attention, providing reliable guidance for constructing accurate sparse attention.
Experimental results show that our method outperforms existing sparse attention approaches in video generation quality and achieves up to 1.75$\times$ end-to-end speedup on GPUs.
Code: \textcolor{blue}{\url{https://github.com/shawnricecake/draft-attention}}

\end{abstract}

\section{Introduction}

Diffusion Transformers (DiTs)~\cite{dit} have emerged as a powerful paradigm for visual generative tasks across both image and video generation, surpassing the traditional UNets~\cite{unet}. 
\begin{wrapfigure}{r}{0.4\linewidth}
\vspace{-3.5mm}
  \centering
  \includegraphics[width=\linewidth]{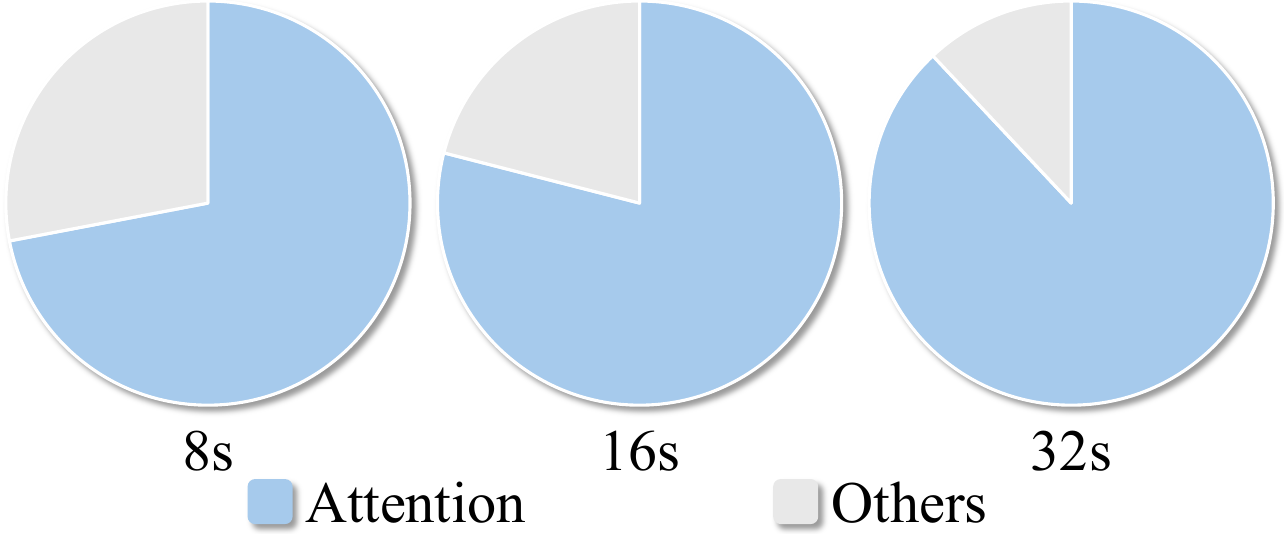}
  \caption{
    FLOPs breakdown for 720p video generation with Hunyuan Video. 
  }
  \label{fig:latency_profiling}
\vspace{-4mm}
\end{wrapfigure}
Video generation with DiTs adopts spatiotemporal 3D full attention to extend image-based generation to the temporal domain~\cite{arnab2021vivit}, leading to   visually coherent high-quality video generation performance~\cite{yang2024cogvideox, kong2024hunyuanvideo, wan2025}, validating the effectiveness of DiTs for video generation. 
Despite the superior generation performance with DiTs, it  remains computationally expensive  due to the attention mechanism in transformers. 
The quadratic complexity with respect to context length~\cite{dao2022flashattention} becomes a significant computational bottleneck when handling sequences with hundreds of thousands of tokens.
For example, as shown in Figure~\ref{fig:latency_profiling}, the Hunyuan Video model~\cite{kong2024hunyuanvideo} spends over 80\% of its total computation on the attention mechanism when generating videos longer than 16 seconds.
As a result, the slow generation speed limits the 
application and deployment of these promising video generation models across a range of practical tasks.

Fortunately, pioneering works~\cite{zhang2023ho, tang2024quest, xiao2023streamingllm, jiang2024minference} on Large Language Models (LLMs)~\cite{gpt2, llama1, llama2, llama3} has demonstrated substantial redundancy in the attention mechanism, offering an opportunity for acceleration by introducing sparsity into the attention.  
Inspired by this, 
recent works~\cite{xi2025sparse_efficient_dit_video_sparseattn, xia2025training} explore the sparse attention methods for video generation models, demonstrating promising speedups while preserving generation quality. 
Specifically, two static sparse attention patterns (targeting spatial and temporal dimensions respectively) are explored in Sparse VideoGen~\cite{xi2025sparse_efficient_dit_video_sparseattn} to reduce redundancy, with relatively significant performance degradation under large sparsity because of non-adaptive static patterns. To mitigate this issue, dynamic sparse attention is investigated in AdaSpa~\cite{xia2025training} to perform full attention once for different prompts as a warm-up to guide subsequent sparsity. 
Although AdaSpa provides prompt-dependent sparse patterns, patterns still remains static during the diffusion process.

\begin{figure*}[t]
  \centering
  \includegraphics[width=1.0\linewidth]{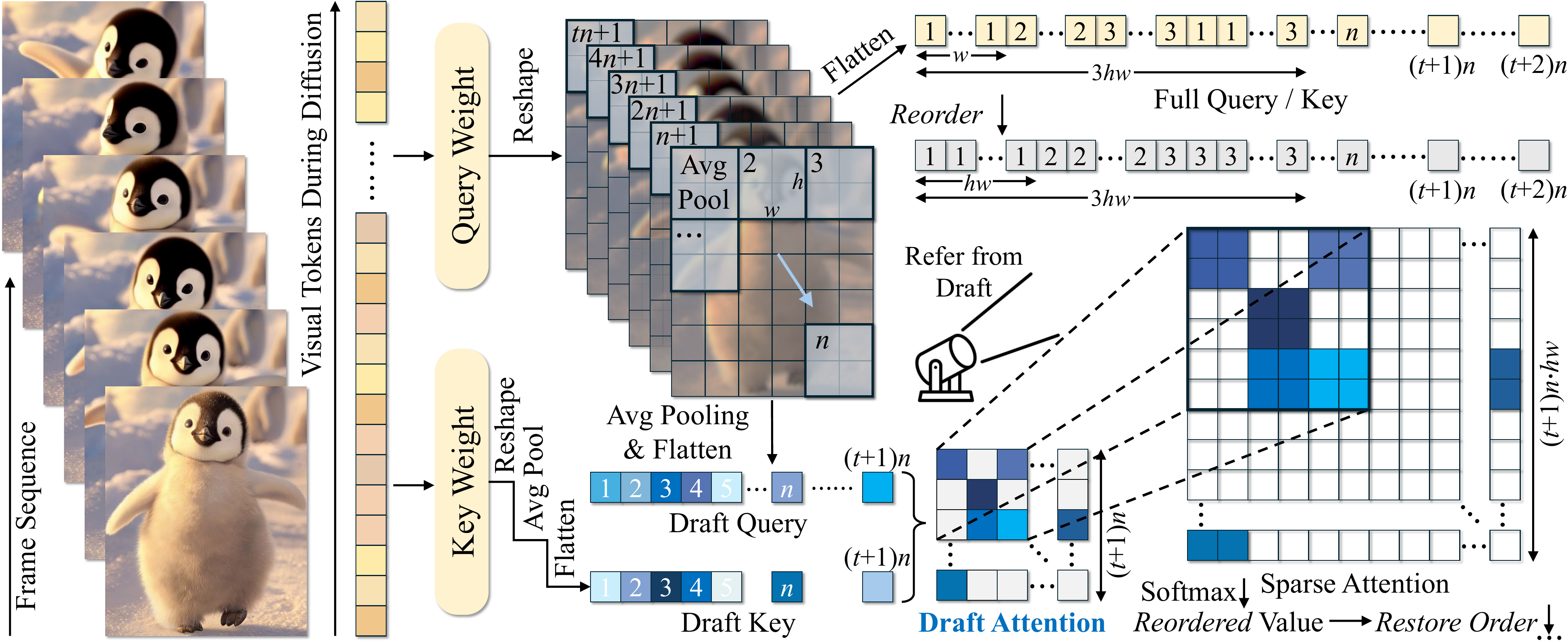}
  \caption{
  Whole \texttt{DraftAttention} Pipeline. Both the query and key are reshaped into sequences of feature maps across frames, then downsampled via average pooling to produce the low-resolution draft query and draft key. 
  Draft attention is computed using the flattened draft query and key.
  The full-resolution query and key need to be reordered for the alignment of draft attention guidance.
  }
  \label{fig:whole_pipeline}
\end{figure*}

\textbf{Framework Overview.}
Motivated by the absence of true dynamic sparse attention at the per-module level, we investigate a more fine-grained design—adapting the sparse attention patterns dynamically for each specific attention module.
In this paper, we propose an efficient sparse attention method, \texttt{DraftAttention}, as shown in Figure~\ref{fig:whole_pipeline}, which leverages draft attention to dynamically generate a sparse pattern for each attention module, enabling efficient acceleration of video diffusion transformers.
The key idea is to compute the draft attention based on downsampled low-resolution  query and key, thus identifying the most important areas in the  attention map with minor computational overhead. 
The resulting low-resolution sparse mask then guides full-resolution sparse attention, with effective reordering applied to ensure fast, hardware-friendly execution.

\textbf{Great Advantages.}
We highlight the following advantages with our draft attention method: \textbf{(i)} ({Efficiency}) The computation of draft attention map is lightweight, as it operates on a reduced number of tokens, thereby lowering the quadratic complexity of the attention mechanism.
\textbf{(ii)} ({Effectiveness}) The draft attention captures high-level representations and preserves essential visual patterns for videos, leading to an effective mask to identify the   critical structures in attention mechanism.
\textbf{(iii)} ({Plug-and-Play}) Our method requires no additional training and integrates seamlessly as a plug-and-play module into existing video diffusion transformers for handling long input sequences.


\textbf{Theoretical Justification.} 
We also present the theoretical analysis that formally characterizes how the low-resolution draft attention effectively guides the full-resolution attention mechanism. 
Specifically, we show that the upper bound of the difference between the full-resolution attention map and the draft attention map remains controlled. 
Meanwhile, we show that the error introduced by the sparse pattern derived from the draft attention map remains bounded.

\textbf{Hardware Friendliness.} 
To align the  region-level sparsity with  token-level computations, we apply a deterministic reordering of 
tokens such that entries in each region  become contiguous in   memory, ensuring hardware-friendly execution of sparse attention.

\textbf{Comprehensive Experiments.} 
In our experiments, we use an 8$\times$16 pooling kernel with a stride equal to the kernel size, reducing the number of tokens by a factor of 128. 
This configuration also matches the efficient block size supported by efficient attention computation frameworks~\cite{dao2022flashattention, guo2024blocksparse}.
Meanwhile, through reordering, we group the scattered sparse patterns into a contiguous format, allowing 128 visual tokens within each kernel to be processed in a single stage—either computed or skipped. This enables both accurate and faster sparse attention at full resolution.
Such aggressive downsampling also incurs minimal computational overhead for the low-resolution draft attention.
Meanwhile, our method outperforms other sparse attention methods on video generation tasks across various resolutions under the same computational budget. It achieves up to a 1.75$\times$ end-to-end speedup on GPUs, demonstrating strong practical efficiency and scalability for long video sequences without compromising generation quality.
Our contributions are summarized as follows,

\begin{itemize}[label={}, leftmargin=*]
\vspace{-2mm}
\item \textbf{1.} We introduce a vision-centric perspective on spatial and temporal redundancy in video diffusion, using pooling to extract high-level representations with a broader receptive field. Building on this, we propose \texttt{DraftAttention}, a hardware-friendly approach that accelerates video diffusion transformers using guidance from low-resolution draft attention.

\item \textbf{2.} We provide a theoretical analysis demonstrating the controlled difference between full-resolution attention and low-resolution draft attention, as well as the bounded error introduced by the sparse pattern derived from the draft attention map, thereby justifying the effectiveness of our design.

\item \textbf{3.} Experimental results show that \texttt{DraftAttention} achieves better video generation quality compared to other sparse attention methods with same computation cost. Meanwhile, on GPUs, our method achieves up to 1.75$\times$ end-to-end acceleration for video generation.

\end{itemize}

\section{Related Works}

\subsection{Efficient Diffusion Models}

\textbf{Diffusion Model Compression.} 
Weight quantization is a common approach to compress diffusion models and achieve acceleration~\cite{li2023qdiffusion}. 
Previous works~\cite{zhang2025sageattention, zhang2024sageattention2, li2024svdquant} propose optimal quantization methods to quantize attention weights to INT8, INT4/FP8, or even FP4, which achieve high compression ratios for the diffusion model size.
Also, other works explore efficient architectures~\cite{xie2025sana} including linear attention or high-compression auto-encoders~\cite{chen2025deep} to accelerate the diffusion and improve model performance, which extends the scalability of diffusion models.
Our method is orthogonal to these techniques and integrates with them to yield additional performance gains.

\textbf{Reduce Diffusion Steps.} 
Some distillation-based works~\cite{li2023snapfusion,yin2024improved} adopt training for the simpler to build few-step diffusion models, which accelerates the diffusion progress by reducing the steps.
However, such distillation techniques require expensive re-training or fine-tuning, which is impractical for the application of most video diffusion models. 
In contrast, our approach directly uses off-the-shelf pre-trained models without any additional training.

\subsection{Sparse Attention Methods}

Attention mechanisms exhibit inherent sparsity~\cite{child2019generating}, allowing computational acceleration by limiting interactions to a subset of the key-value pair. 
StreamingLLM~\cite{xiao2023streamingllm} explores the temporal locality with attention sinks to further preserve sparse attention model performance. 
H2O~\cite{zhang2023ho} identifies a small set of Heavy Hitter tokens that dominate overall attention scores. DuoAttention~\cite{xiao2025duoattention} and MInference~\cite{jiang2024minference} demonstrate distinct sparse patterns across different attention heads.
XAttention~\cite{xu2025xattention_efficient_sparseattn} leverages the sum of antidiagonal values in the attention matrix to provide a powerful proxy for block importance,
resulting in high sparsity and dramatically accelerated inference.
Sparse VideoGen~\cite{xi2025sparse_efficient_dit_video_sparseattn} explores spatial and temporal heads in video diffusion models to improve the inference
efficiency.
AdaSpa~\cite{xia2025training} applies dynamic block-sparse masking with online token importance search, accelerating video diffusion without fine-tuning.
These works collectively show that such transformer-based models 
contain significant redundancy in their attention mechanisms.
This motivates our exploration of dynamic, fine-grained sparse attention patterns for video diffusion transformers.


\section{Methodology}

We  introduce the framework of our draft attention in great detail to first identify critical areas in draft attention with a low-resolution mask and then apply the mask to full-resolution attention. 
Next theoretical analysis for the draft attention and the corresponding sparse attention is presented to demonstrate the
effectiveness of our design.
Moreover, we provide a deterministic reordering of tokens to align the region-level sparsity with token-level computation,  ensuring efficient hardware-friendly execution.


\subsection{Draft Attention}\label{sec:draft}

Full attention over long video sequences is prohibitively expensive due to its quadratic complexity in sequence length. However, many interactions in video are spatially and temporally localized. We leverage this structure by introducing a two-stage attention mechanism: a lightweight \emph{draft attention} phase that estimates regional relevance, followed by a masked sparse attention applied to the full-resolution sequence.

We first define the full attention computation below.

\begin{definition}[Full Attention]\label{def:attn_full}
Given hidden states $X \in \mathbb{R}^{n \times d}$, the full attention output is:
\begin{align}
\mathsf{Attn}(X) = \mathsf{Softmax}\left( \frac{Q K^\top}{\sqrt{d}} \right) V \in \mathbb{R}^{n \times d},    
\end{align}
where $Q = X W_Q$, $K = X W_K$, $V = X W_V$ are the query, key, and value projections, and $W_Q, W_K, W_V \in \mathbb{R}^{d \times d}$ are learned weight matrices.
\end{definition}

To reduce computation, we downsample $Q$ and $K$ via average pooling, forming a low-resolution draft attention map to guide sparsity.

\begin{definition}[Draft Attention via Average Pooling]\label{def:attn_draft}
Given hidden states $X \in \mathbb{R}^{n \times d}$, representing spatial-temporal tokens across frames, we partition the sequence into $g \ll n$ disjoint regions $\{R_i\}_{i=1}^g$, where each region $R_i \subset [n]$ corresponds to a pooled spatial patch over time. 
Each $R_i$ is an unordered set of token indices.
Let $Q$ and $K$ be the projected queries and keys. The draft query and draft key representations are obtained by average pooling over each region:
\begin{align}
\tilde{Q}_i = \frac{1}{|R_i|} \sum_{j \in R_i} Q_j, \quad
\tilde{K}_i = \frac{1}{|R_i|} \sum_{j \in R_i} K_j, \quad \text{for } i = 1, \dots, g.    
\end{align}

The resulting low-resolution \textbf{draft attention map} is computed as:
\begin{align}
A_{\mathrm{draft}} = \mathsf{Softmax}\left( \frac{\tilde{Q} \tilde{K}^\top}{\sqrt{d}} \right) \in \mathbb{R}^{g \times g}.    
\end{align}

This map approximates region-level relevance and is used to guide sparse attention over the full-resolution sequence.
\end{definition}

The computation cost of the low-resolution draft attention map is minor compared with the full-resolution attention computation, as it operates on a reduced 
number of tokens  and thereby lowers the quadratic complexity of the attention mechanism.

\paragraph{Guided Sparsity via Draft Attention.}

To reduce the cost of full attention, we extract a structured sparsity pattern from the draft attention map $A_{\mathrm{draft}} \in \mathbb{R}^{g \times g}$ by retaining only a fraction $r \in (0,1)$ of the most salient region-to-region interactions. We define a binary mask $M \in \{0,1\}^{g \times g}$, where $M_{ij} = 1$ indicates that region $R_i$ is permitted to attend to region $R_j$, and $M_{ij} = 0$ otherwise. The mask is constructed by selecting the top-scoring entries in $A_{\mathrm{draft}}$ under a fixed sparsity ratio $r$.

To align the region-level sparsity with token-level computation, we apply a deterministic reordering of tokens such that entries in each region $R_i$ become contiguous. This facilitates efficient masking and block-wise computation in sparse attention. We provide more details for reordering in Section~\ref{sec:reorder}.

This region-level sparsity pattern is then lifted to token resolution by defining a full-resolution binary mask $\widehat{M} \in \{0,1\}^{n \times n}$:
\begin{align}
\widehat{M}_{uv} = M_{ij} \quad \text{if } u \in R_i,\; v \in R_j.
\end{align}
In general, the attention map is split into multiple non-overlapping regions by the pooling kernels. For each region, all its elements are either computed for attention or skipped for acceleration. The determination for whether to skip each region is denoted by the low-resolution binary mask $M$ for all regions, with $\widehat{M}$ as its full-resolution mask for all elements (i.e., tokens).

Sparse attention is then computed by applying the mask to the full attention scores:
\begin{align}
\mathsf{SparseAttn}(X) = \mathsf{Softmax}\left( \left( \frac{QK^\top}{\sqrt{d}} \right) \odot \widehat{M} \right) V,    
\end{align}
where $\odot$ denotes element-wise/Hadamard product. This formulation retains the most relevant interactions while enforcing structured sparsity for improved computational efficiency.

\subsection{Theoretical Analysis}
\label{sec:analysis}

We present Frobenius-norm bounds quantifying the error introduced by our two-stage approximation strategy:
(1) average pooling (draft attention), and
(2) structured sparsification via top-$r$ indexing.

\subsubsection{Error from Draft Attention}

Let the input sequence be partitioned into $g$ disjoint regions $\{R_i\}_{i=1}^g$ of equal size $|R_i| = n/g$. Define the full-resolution attention logits and their pooled approximation as:
\begin{align}
  S_{uv} := \langle Q_u, K_v \rangle,
  \quad
  \widetilde S_{ij} := \langle \widetilde Q_i, \widetilde K_j \rangle,
  \qquad u,v \in [n],\; i,j \in [g],   
\end{align}
where $\widetilde Q_i = \frac{1}{|R_i|} \sum_{u \in R_i} Q_u$ and similarly for $\widetilde K_j$.

We restore the region-level scores $\widetilde S \in \mathbb{R}^{g \times g}$ to full resolution by defining a block-constant approximation:
\begin{align}
  (S_{\mathrm{draft}})_{uv} := \widetilde S_{ij}
  \quad \text{for } u \in R_i,\; v \in R_j.
\end{align}

Define the worst-case deviation between token-level logits and their region-averaged counterpart as:
\begin{align}
  \delta := \max_{i,j} \max_{u \in R_i,\, v \in R_j} \big| S_{uv} - \widetilde S_{ij} \big|.
\end{align}

\begin{theorem}[Draft Attention Error]
\label{thm:draft_error}
If all regions have equal size $|R_i| = n/g$, then the Frobenius-norm error between the full and draft logit matrices is bounded by:
\begin{align}
  \| S - S_{\mathrm{draft}} \|_F \le \delta\,n.
\end{align}
\end{theorem}

The detailed proof is shown in Appendix~\ref{sec:app:proofs}.

\begin{remark}
Theorem~\ref{thm:draft_error} quantifies the approximation error introduced by replacing token-level attention logits with block-wise averages obtained via average pooling. In practice, if tokens within a region are similar—such as in videos with local temporal consistency or spatial smoothness—the difference $|S_{uv} - \widetilde S_{ij}|$ remains small for most $(u,v)$. Consequently, the overall Frobenius-norm error $\|S - S_{\mathrm{draft}}\|_F$ scales with a modest $\delta$, leading to minimal distortion in the attention structure. This justifies using the low-resolution draft map as a proxy for full-resolution attention in computationally constrained settings.
\end{remark}

\subsubsection{Error from Sparsity Mask}

We now consider the additional error introduced by sparsifying the logits based on the top-$r$ draft attention values. Let $\widetilde S_{(1)} \ge \cdots \ge \widetilde S_{(g^2)}$ be the sorted region-level scores. Define the threshold:
\begin{align}
  t := \widetilde S_{(\lceil r g^2 \rceil)},
\end{align}
and let $M_{ij} = 1$ if $\widetilde S_{ij} \ge t$ and $0$ otherwise. The mask is lifted to token resolution by $\widehat M_{uv} = M_{ij}$ for $u \in R_i,\; v \in R_j$.

\begin{theorem}[Sparsity Mask Error]
\label{thm:mask_error}
Under uniform region size $|R_i| = n/g$, the error from masking the logits satisfies:
\begin{align}
  \| S - S \odot \widehat M \|_F \le n(\delta + t) \sqrt{1 - r}.
\end{align}
\end{theorem}

The detailed proof is shown in Appendix~\ref{sec:app:proofs}.

\begin{remark}
Theorem~\ref{thm:mask_error} captures the additional error introduced by enforcing structured sparsity through top-$r$ indexing. This error is controlled by the threshold value $t$, which defines the weakest retained region-level interaction. When the draft attention distribution is concentrated—i.e., a small fraction of regions account for most of the mass—top-$r$ masking retains the most informative blocks while discarding only low-scoring interactions, keeping $(\delta + t)$ small. This ensures that sparsification preserves the dominant attention patterns with bounded deviation.
\end{remark}

Together, Theorems~\ref{thm:draft_error} and~\ref{thm:mask_error} provide a principled decomposition of the total approximation error: one from average pooling, and one from sparsity. Their combined bound shows that draft attention is an efficient surrogate for full attention, maintaining structural fidelity while enabling substantial computational savings. This justifies its use in long-context video diffusion transformers, where local smoothness and sparse relevance patterns are common.








\begin{figure*}[t]
  \centering
  \includegraphics[width=1.0\linewidth]{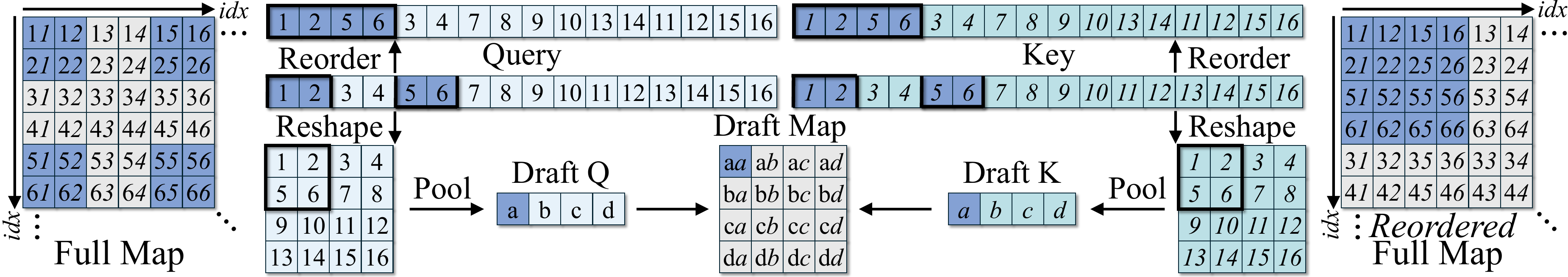}
  \caption{
  Illustration for the necessity of the reordering. 
  The "$\text{x}y$" in attention map denotes attentivity between token $\text{x}$ in query and token $y$ in key.
  Grouping the sparse pattern enables hardware-friendly layout, leading to faster attention computation.
  }
  \label{fig:reorder_explain}
\end{figure*}

\subsection{Reordering for Patch-Aligned Sparse Attention}\label{sec:reorder}

\begin{wrapfigure}{r}{0.53\textwidth}
\vspace{-1mm}
\vspace{-1em}
\begin{minipage}{\linewidth}
\begin{algorithm}[H]
\caption{Generate Reorder Index}
\label{alg:reorder}
\KwIn{Frame size $(H, W)$, patch size $(h, w)$, number of frames $F$}
\KwOut{Permutation $\pi \in [n]$ where $n = F \cdot H \cdot W$}

$\pi \gets []$\;

\For{$f = 0$ \KwTo $F - 1$}{
  \For{$i = 0$ \KwTo $H/h - 1$}{
    \For{$j = 0$ \KwTo $W/w - 1$}{
      \For{$u = 0$ \KwTo $h - 1$}{
        \For{$v = 0$ \KwTo $w - 1$}{
          $y \gets i \cdot h + u$, $x \gets j \cdot w + v$\;
          ${\tt idx} \gets f \cdot H \cdot W + y \cdot W + x$\;
          Append ${\tt idx}$ to $\pi$\;
        }
      }
    }
  }
}
\Return{$\pi$}
\end{algorithm}
\end{minipage}
\vspace{-2em}
\vspace{-3mm}
\end{wrapfigure}

To enable accurate and efficient sparse attention that respects spatial structure, we apply a deterministic reordering algorithm (Algorithm~\ref{alg:reorder}) to the flattened full-resolution token sequence. As shown in Figure~\ref{fig:reorder_explain}, the goal is to align the memory layout of full-resolution tokens with the spatial region structure used in low-resolution draft attention. This alignment ensures that the region-level sparsity patterns are directly and efficiently propagated to full-resolution attention through block-level masking.


\paragraph{Justification.}
In the default row-major layout, spatial tokens are appended row-wise within each frame, causing spatial patches to be scattered in memory. This fragmentation hinders efficient usage of sparse attention kernels, which rely on contiguous blocks in fixed size for the optimal performance.
As illustrated in Figure~\ref{fig:reorder_explain}, tokens 1, 2, 5, and 6 are spatial neighbors but are not stored consecutively in the memory of full attention map (i.e., left side of Figure~\ref{fig:reorder_explain}) due to the presence of tokens 3 and 4. 
While it is still possible to gather these tokens and compute their average, this process is highly inefficient. 
Similarly, masking out these scattered blocks is also inefficient, as it effectively reduces the block size, which in turn lowers arithmetic intensity, causes uncoalesced memory access, and increases the number of kernel launches.

\paragraph{Design.}
We divide each frame into non-overlapping patches of size $h \times w$. For each frame, tokens within the same patch are grouped contiguously. Unlike prior methods (e.g., SVG~\cite{xi2025sparse_efficient_dit_video_sparseattn}) that overlook misalignment issues when the kernel size does not divide evenly into the latent feature map size, our per-frame design preserves the completeness of each feature map, generating  more reliable captured high-level representations.
Meanwhile, this per-frame design ensures that each patch in a frame is stored as a contiguous block, matching the structure of the downsampled low-resolution queries and keys used in draft attention.
For instance, tokens 1, 2, 5, and 6 belong to the same patch and are reordered to appear consecutively in both the query and key sequences, as illustrated at the top of Figure~\ref{fig:reorder_explain}. This reordering ensures that each entry in draft attention map (e.g., a\textit{a}) corresponds to a specific block ($\{1,2,5,6\}$ from query and $\{\textit{1},\textit{2},\textit{5},\textit{6}\}$ from key) within reordered full attention map.

\paragraph{Execution.}
Applying the permutation $\pi$ ensures that tokens grouped in each $h \times w$ patch are stored contiguously in memory, enabling efficient block-wise indexing and masking. 
This structured layout aligns the memory access pattern with the computational needs of sparse attention operations.
This is especially critical for efficient execution with frameworks like FlashAttention~\cite{dao2022flashattention} and Block Sparse Attention~\cite{guo2024blocksparse}, which leverage fused GPU kernels that operate on fixed-size blocks.


\paragraph{Restoration.}
After sparse attention is applied in the reordered space (i.e., the attention computation for reordered query, key, and value), we apply the inverse permutation $\pi^{-1}$ (Algorithm~\ref{alg:restore}) to restore the original spatial-temporal layout for the following correct model inference.

\begin{wrapfigure}{r}{0.44\textwidth}
\vspace{-1em}
\vspace{-1.1mm}
\begin{minipage}{\linewidth}
\begin{algorithm}[H]
\caption{Generate Restore Index}
\label{alg:restore}
\KwIn{Permutation $\pi \in [n]$}
\KwOut{Inverse permutation $\pi^{-1}$}

Initialize $\pi^{-1} \gets$ zero array of length $n$\;

\For{$i = 0$ \KwTo $n - 1$}{
  $\pi^{-1}_{\pi_i} \gets i$\;
}
\Return{$\pi^{-1}$}
\end{algorithm}
\end{minipage}
\vspace{-1em}
\end{wrapfigure}

\paragraph{Benefit.}
This reordering bridges the gap between the coarse-grained sparsity structure derived from draft attention and the fine-grained full-resolution attention computation. 
This layout guarantees that pooled regions align cleanly with memory blocks, preserving spatial locality and enabling predictable, coalesced memory access.
As a result, it supports efficient masking and ensures compatibility with high-throughput attention kernels. This design significantly reduces overhead and maximizes hardware efficiency during attention computation.

\section{Experimental Results}

\subsection{Experiment Setup}

\paragraph{Model Family.}
We  adopt open-sourced state-of-the-art
video generation models in our experiments, including HunyuanVideo-T2V~\cite{kong2024hunyuanvideo} for 768p resolution with 128 frames and Wan2.1-T2V~\cite{wan2025} for both 512p and 768p resolutions with 80 frames.
We use 512p and 768p resolutions to align with the 8$\times$16 average pooling kernel (with stride equal to the kernel size), enabling convenient and consistent downsampling of visual tokens during the diffusion process.
This is because the corresponding latent sizes—32$\times$48 for 512p and 48$\times$80 for 768p—are perfectly divisible by the 8$\times$16 kernel, ensuring efficient and artifact-free pooling.
Note that our method supports video generation at any resolution by applying appropriate padding.
Following prior works~\cite{xi2025sparse_efficient_dit_video_sparseattn, li2023distrifusion, li2024svdqunat, liu2024timestep}, we retain full attention across all methods for the first 25\% of denoising steps to preserve the video generation quality.
We adopt Block Sparse Attention~\cite{guo2024blocksparse} for the implementation of our method and mainly compare our method with the Sparse VideoGen (SVG)~\cite{xi2025sparse_efficient_dit_video_sparseattn}.
We observe discrepancies in the generation results of the Wan2.1-T2V model between our method and SVG, due to difference of codebases. To ensure a fair comparison, we provide results using full attention for both methods.

\paragraph{Metrics and Prompts.}
We evaluate the quality of generated videos with VBench~\cite{huang2023vbench}, and the similarity of   generated videos with metrics including  Peak Signal-to-Noise Ratio (PSNR), Structural Similarity Index Measure (SSIM), and Learned Perceptual Image Patch Similarity (LPIPS)~\cite{zhang2018perceptual}.
Especially, we report the image quality, subject consistency, background consistency, dynamic degree, and aesthetic quality from VBench for our generated videos.
All videos are generated with the prompts from the Penguin Video Benchmark~\cite{kong2024hunyuanvideo} released by HunyuanVideo.
The reported computation cost in PFLOPs includes the main diffusion transformer models, and the latency results are all tested on the H100 GPU.

\begin{table}[]
\centering
\caption{
Main results of the proposed method compared to the Sparse VideoGen (SVG)~\cite{xi2025sparse_efficient_dit_video_sparseattn}.
}
\label{tab:main_results}
\resizebox{1.0\linewidth}{!}{
\begin{tabular}{c|c|c|ccc|ccccc|c}
\toprule
\multirow{2}{*}{Model}                                                    & \multirow{2}{*}{Method} & Sparse & PSNR  & SSIM  & LPIPS & Img.   & Sub.   & Bakg.  & Dyn.   & Aes.   & PFLOPs \\
 &                         & Ratio    & $\uparrow$    & $\uparrow$    & $\downarrow$    & Qual.  & Cons.  & Cons.  & Deg.   & Qual.  & $\downarrow$     \\
 \midrule
\multirow{6}{*}{\begin{tabular}[c]{@{}c@{}}Wan2.1\\ (512p)\end{tabular}}  & \multirow{3}{*}{SVG}    & 0\%      & /     & /     & /     & 65.1\% & 95.0\% & 95.9\% & 44.7\% & 58.9\% & 145.65   \\
\cline{3-12}
 &                         & 55\%     & 25.61 & 83.63 & 10.42 & 65.2\% & 94.8\% & 95.9\% & 45.2\% & 58.9\% & 99.26   \\
 &                         & 75\%     & 23.66 & 78.80 & 15.05 & 64.7\% & 94.5\% & 95.7\% & 45.7\% & 58.6\% & 91.12   \\
\cline{2-12}
 & \multirow{3}{*}{Ours}   & 0\%      & /     & /     & /     & 69.3\% & 95.5\% & 96.7\% & 47.6\% & 61.5\% & 145.65   \\
\cline{3-12}
 &                         & 55\%     & 25.13 & \textbf{84.77} & \textbf{8.43}  & 69.2\% & 95.5\% & 96.6\% & 47.6\% & 61.5\% & 99.26   \\
 &                         & 75\%     & 23.10 & \textbf{79.07} & \textbf{12.37} & 69.0\% & 95.4\% & 96.5\% & 46.9\% & 61.5\% & 91.12   \\
\midrule
\multirow{6}{*}{\begin{tabular}[c]{@{}c@{}}Wan2.1\\ (768p)\end{tabular}}  & \multirow{3}{*}{SVG}    & 0\%      & /     & /     & /     & 67.7\% & 95.3\% & 96.4\% & 43.4\% & 60.4\% & 609.52   \\
\cline{3-12}
 &                         & 55\%     & 26.01 & 84.81 & 10.89  & 67.9\% & 95.1\% & 96.3\% & 42.1\% & 60.0\%  &  354.68   \\
 &                         & 75\%     & 23.62 & 79.05 & 17.57  & 67.5\% & 94.8\% & 96.1\% & 42.1\% & 58.8\% & 309.95   \\
\cline{2-12}
 & \multirow{3}{*}{Ours}   & 0\%      & /     & /     & /     & 67.5\% & 95.7\% & 97.1\% & 37.7\% & 60.8\% & 609.52   \\
 \cline{3-12}
 &                         & 55\%     & \textbf{29.22} & \textbf{92.16} & \textbf{5.82}  & 67.4\% & 95.6\% & 97.0\% & 37.2\% & 60.8\% & 354.69   \\
 &                         & 75\%     & \textbf{27.17} & \textbf{88.97} & \textbf{8.71}  & 67.2\% & 95.6\% & 97.0\% & 38.6\% & 60.7\% & 309.95   \\
 \midrule
\multirow{7}{*}{\begin{tabular}[c]{@{}c@{}}Hunyuan\\ (768p)\end{tabular}} & Dense                   & 0\%      & /     & /     & /     & 66.4\% & 96.0\% & 97.0\% & 36.4\% & 58.6\% & 682.67  \\
\cline{2-12}
 & \multirow{3}{*}{SVG}    & 60\%     & 25.80 & 84.46 & 14.20 & 66.4\% & 95.9\% & 97.0\% & 36.6\% & 58.2\% & 343.72  \\
 &                         & 80\%     & 24.70 & 81.90 & 17.55 & 66.0\% & 95.7\% & 96.9\% & 33.9\% & 58.1\% & 295.30  \\
 &                         & 90\%     & 23.48 & 78.57 & 22.60 & 65.1\% & 95.4\% & 96.7\% & 32.8\% & 57.5\% & 283.20  \\
 \cline{2-12}
 & \multirow{3}{*}{Ours}   & 60\%     & \textbf{32.08} & \textbf{93.21} & \textbf{5.58}  & \textbf{66.4\%} & \textbf{95.9\%} & \textbf{97.0\%} & 35.9\% & \textbf{58.5\%} & 343.73  \\
 &                         & 80\%     & \textbf{29.19} & \textbf{89.32} & \textbf{9.19}  & \textbf{66.2\%} & \textbf{95.8\%} & \textbf{97.0\%} & \textbf{35.7\%} & \textbf{58.2\%} & 295.31  \\
 &                         & 90\%     & \textbf{24.22} & \textbf{79.90} & \textbf{18.12} & \textbf{65.9\%} & \textbf{95.7\%} & \textbf{96.9\%} & \textbf{36.6\%} & \textbf{57.8\%} & 283.20  \\
 \bottomrule
\end{tabular}
}
\end{table}

\subsection{Main Results}

\textbf{Higher Generation Quality.}
We provide the main results compare with the SVG method in Table~\ref{tab:main_results}.
To perform a comprehensive study, different sparsity ratios for the attention mechanism are evaluated under various resolutions with multiple video generation model architectures. 
With the Wan2.1 model, we observe that our method achieves less image quality degradation compared with SVG. 
The similarity results  measured by PSNR, SSIM and LPIPS demonstrate that our method generates videos more similar to the dense model compared with SVG under the same sparsity. Specifically, for Wan2.1 (768p), our method achieves non-marginal improvements over SVG  on PSNR, SSIM and LPIPS (such as our 8.71 LPIPS \textit{v.s.} 17.57 LPIPS from SVG under 75\% sparsity). 
For  the Hunyuan model, our method achieves  better performance across almost all reported metrics, under a fair comparison with SVG following the same sparsity and computational cost in PFLOPs.
Although SVG includes additional overhead for spatial or temporal head selection,  we exclude this computation cost from the reported PFLOPSs of SVG in Table~\ref{tab:main_results}.
Note that the additional overhead of our \texttt{DraftAttention} is minor, leading to almost the same computations as SVG in the table.

\begin{figure*}[t]
  \vspace{-2mm}
  \centering
  \includegraphics[width=1.0\linewidth]{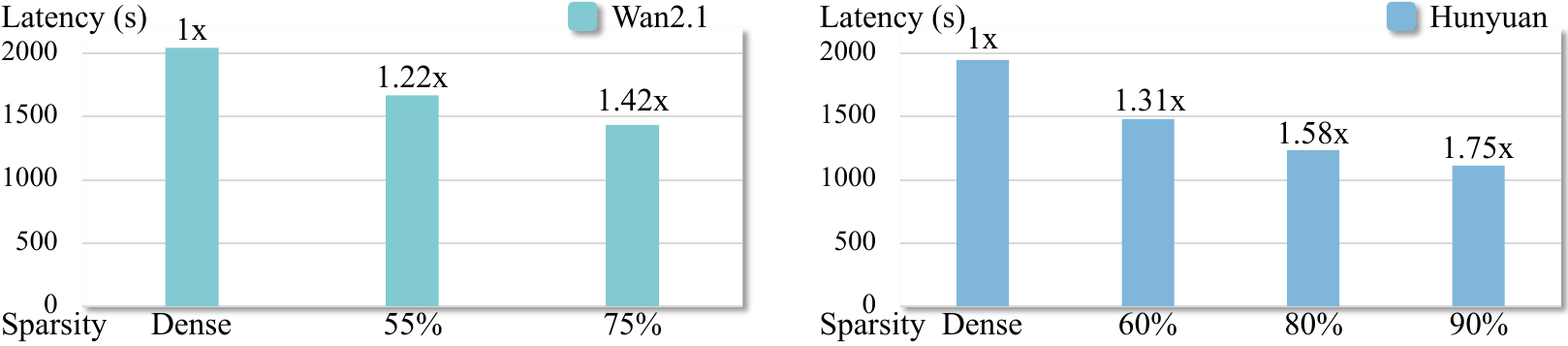}
  \vspace{-5mm}
  \caption{
  Latency results tested in 768p with H100 GPU for different sparsity ratios in attention.
  }
  \label{fig:latency_results}
  \vspace{-3mm}
\end{figure*}

\textbf{Superior Inference Acceleration.}
Furthermore, we provide our latency results  in Figure~\ref{fig:latency_results}.
The latency results are tested on H100 for both Huyuan and Wan2.1 models in 768p resolution.
Our method achieves over 1.75$\times$ acceleration on an H100 GPU with 90\% sparsity in the attention mechanism—demonstrating our outstanding  practical efficiency. 


\begin{figure*}[t]
  \centering
  \includegraphics[width=1.0\linewidth]{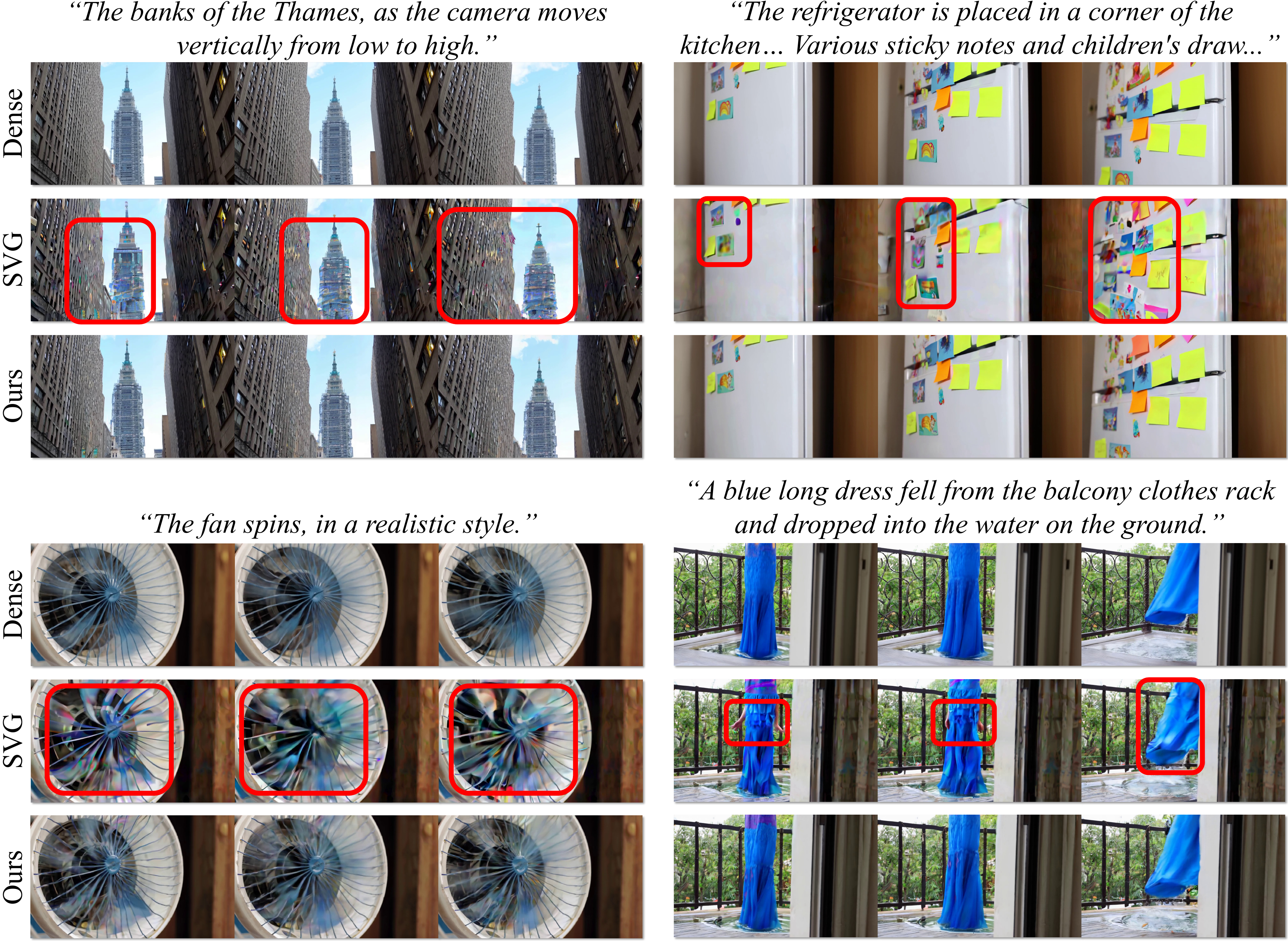}
  \caption{
  Visualization for our method and SVG~\cite{xi2025sparse_efficient_dit_video_sparseattn} with 90\% sparsity ratio in attention.
  }
  \label{fig:visualization}
\end{figure*}

\textbf{Better Visualization.}
We provide the visualization for the comparison between \texttt{DraftAttention} and   SVG   in Figure~\ref{fig:visualization}.
All videos 
are generated with 90\% sparsity in  sparse attention. 
As highlighted in the red box, SVG exhibits a noticeable degradation in generation quality, with apparent  
blurry pixels. 
In contrast, our method better maintains the generation quality with videos more similar to the dense baseline.
We provide generated videos for further visualization comparison in supplementary.

\subsection{Ablation Study}

\begin{figure*}[t]
  \centering
  \includegraphics[width=1.0\linewidth]{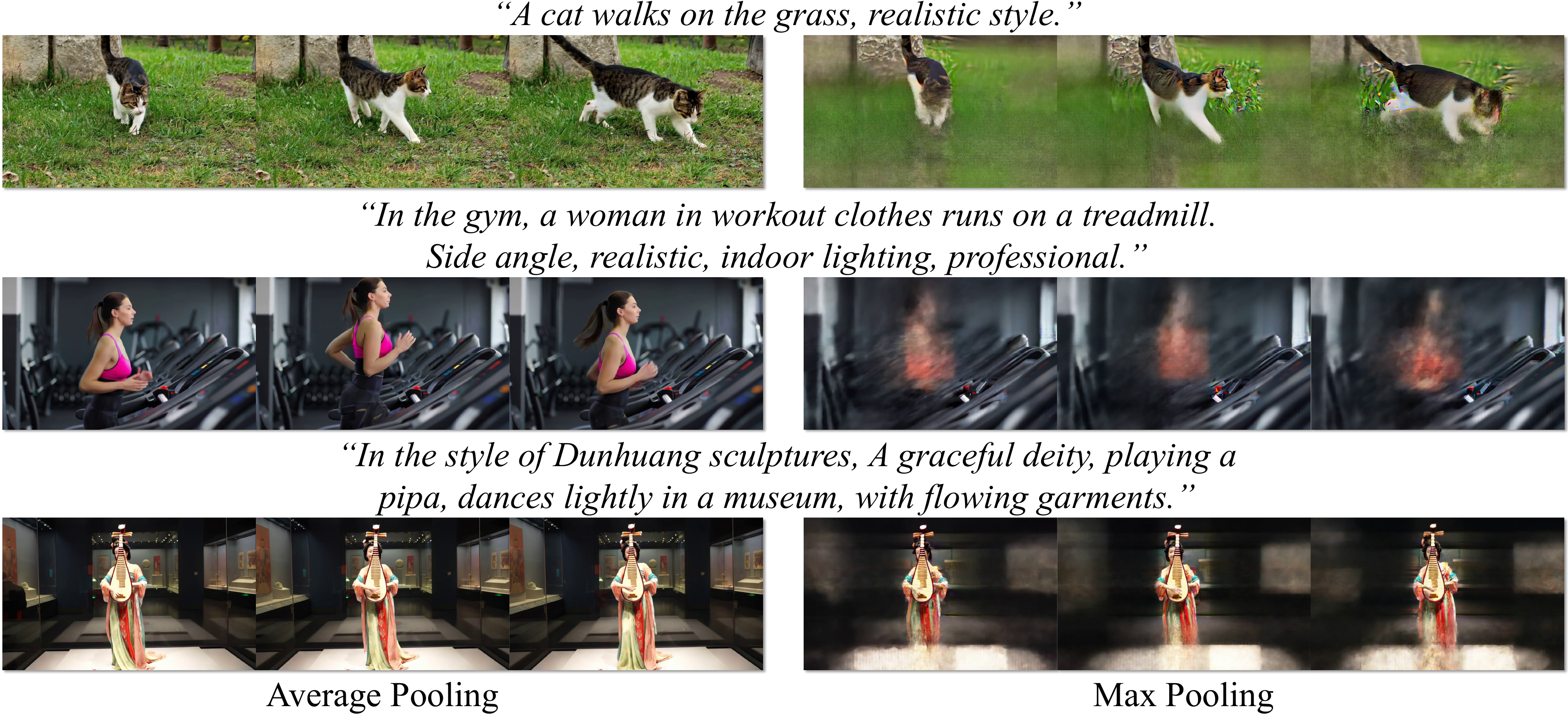}
  \caption{
  Visualization from the ablation study comparing average pooling and max pooling kernels for downsampling in the draft attention module with 90\% sparsity.
  }
  \label{fig:ablation_visualization}
\end{figure*}

As shown in Figure~\ref{fig:ablation_visualization}, we provide the visualization of ablation study for the different downsampling kernels with average pooling and max pooling.
The visualization is generated using 90\% sparsity in the sparse attention, with only the average pooling replaced by max pooling in our framework.
We observe that the average pooling achieves much better generation quality than the max pooling, especially for the background part.

\section{Conclusion}

In this paper, we propose the \texttt{DraftAttention} for the fast video diffusion.
We adopt pooling to compute a low-resolution draft attention map to guide the sparse attention over full-resolution query, key, and value representations.
Combined with effective reordering, this approach achieves fast, hardware-friendly execution on GPUs.
Theoretical analysis is further provided for the justification of our design.
Experiments show that our method outperforms other methods and achieves up to 1.75$\times$ end-to-end acceleration on GPUs.
In the future work, we plan to introduce the quantization for the further acceleration of high-resolution and long-duration video generation on GPUs.

\bibliographystyle{unsrt}
\bibliography{reference}

\begin{thebibliography}{10}

\bibitem{dit}
William Peebles and Saining Xie.
\newblock Scalable diffusion models with transformers.
\newblock {\em arXiv preprint arXiv:2212.09748}, 2022.

\bibitem{unet}
Olaf Ronneberger, Philipp Fischer, and Thomas Brox.
\newblock U-net: Convolutional networks for biomedical image segmentation.
\newblock In {\em Medical image computing and computer-assisted intervention--MICCAI 2015: 18th international conference, Munich, Germany, October 5-9, 2015, proceedings, part III 18}, pages 234--241. Springer, 2015.

\bibitem{arnab2021vivit}
Anurag Arnab, Mostafa Dehghani, Georg Heigold, Chen Sun, Mario Lu{\v{c}}i{\'c}, and Cordelia Schmid.
\newblock Vivit: A video vision transformer.
\newblock In {\em Proceedings of the IEEE/CVF international conference on computer vision}, pages 6836--6846, 2021.

\bibitem{yang2024cogvideox}
Zhuoyi Yang, Jiayan Teng, Wendi Zheng, Ming Ding, , et~al.
\newblock Cogvideox: Text-to-video diffusion models with an expert transformer.
\newblock {\em arXiv preprint arXiv:2408.06072}, 2024.

\bibitem{kong2024hunyuanvideo}
Weijie Kong, Qi~Tian, Zijian Zhang, Rox Min, Zuozhuo Dai, Jin Zhou, Jiangfeng Xiong, Xin Li, et~al.
\newblock Hunyuanvideo: A systematic framework for large video generative models, 2024.

\bibitem{wan2025}
Ang Wang, Baole Ai, Bin Wen, Chaojie Mao, Chen-Wei Xie, Di~Chen, Feiwu Yu, Haiming Zhao, Jianxiao Yang, Jianyuan Zeng, Jiayu Wang, Jingfeng Zhang, Jingren Zhou, Jinkai Wang, Jixuan Chen, Kai Zhu, Kang Zhao, Keyu Yan, Lianghua Huang, Mengyang Feng, Ningyi Zhang, Pandeng Li, Pingyu Wu, Ruihang Chu, Ruili Feng, Shiwei Zhang, Siyang Sun, Tao Fang, Tianxing Wang, Tianyi Gui, Tingyu Weng, Tong Shen, Wei Lin, Wei Wang, Wei Wang, Wenmeng Zhou, Wente Wang, Wenting Shen, Wenyuan Yu, Xianzhong Shi, Xiaoming Huang, Xin Xu, Yan Kou, Yangyu Lv, Yifei Li, Yijing Liu, Yiming Wang, Yingya Zhang, Yitong Huang, Yong Li, You Wu, Yu~Liu, Yulin Pan, Yun Zheng, Yuntao Hong, Yupeng Shi, Yutong Feng, Zeyinzi Jiang, Zhen Han, Zhi-Fan Wu, and Ziyu Liu.
\newblock Wan: Open and advanced large-scale video generative models.
\newblock {\em arXiv preprint arXiv:2503.20314}, 2025.

\bibitem{dao2022flashattention}
Tri Dao, Daniel~Y. Fu, Stefano Ermon, Atri Rudra, and Christopher R{\'e}.
\newblock Flash{A}ttention: Fast and memory-efficient exact attention with {IO}-awareness.
\newblock In {\em Advances in Neural Information Processing Systems (NeurIPS)}, 2022.

\bibitem{zhang2023ho}
Zhenyu Zhang, Ying Sheng, Tianyi Zhou, Tianlong Chen, Lianmin Zheng, Ruisi Cai, Zhao Song, Yuandong Tian, Christopher Re, Clark Barrett, Zhangyang Wang, and Beidi Chen.
\newblock H2o: Heavy-hitter oracle for efficient generative inference of large language models.
\newblock In {\em Thirty-seventh Conference on Neural Information Processing Systems}, 2023.

\bibitem{tang2024quest}
Jiaming Tang, Yilong Zhao, Kan Zhu, Guangxuan Xiao, Baris Kasikci, and Song Han.
\newblock Quest: Query-aware sparsity for efficient long-context llm inference, 2024.

\bibitem{xiao2023streamingllm}
Guangxuan Xiao, Yuandong Tian, Beidi Chen, Song Han, and Mike Lewis.
\newblock Efficient streaming language models with attention sinks.
\newblock {\em arXiv}, 2023.

\bibitem{jiang2024minference}
Huiqiang Jiang, Yucheng Li, Chengruidong Zhang, Qianhui Wu, Xufang Luo, Surin Ahn, Zhenhua Han, Amir~H. Abdi, Dongsheng Li, Chin-Yew Lin, Yuqing Yang, and Lili Qiu.
\newblock {MI}nference 1.0: Accelerating pre-filling for long-context {LLM}s via dynamic sparse attention.
\newblock In {\em The Thirty-eighth Annual Conference on Neural Information Processing Systems}, 2024.

\bibitem{gpt2}
Alec Radford, Jeff Wu, Rewon Child, David Luan, Dario Amodei, and Ilya Sutskever.
\newblock Language models are unsupervised multitask learners.
\newblock 2019.

\bibitem{llama1}
Hugo Touvron, Thibaut Lavril, Gautier Izacard, Xavier Martinet, Marie-Anne Lachaux, Timoth{\'e}e Lacroix, Baptiste Rozi{\`e}re, Naman Goyal, Eric Hambro, Faisal Azhar, et~al.
\newblock Llama: Open and efficient foundation language models.
\newblock {\em arXiv preprint arXiv:2302.13971}, 2023.

\bibitem{llama2}
Hugo Touvron, Louis Martin, Kevin Stone, Peter Albert, Amjad Almahairi, Yasmine Babaei, Nikolay Bashlykov, Soumya Batra, Prajjwal Bhargava, Shruti Bhosale, et~al.
\newblock Llama 2: Open foundation and fine-tuned chat models.
\newblock {\em arXiv preprint arXiv:2307.09288}, 2023.

\bibitem{llama3}
Aaron Grattafiori, Abhimanyu Dubey, Abhinav Jauhri, Abhinav Pandey, Abhishek Kadian, Ahmad Al-Dahle, Aiesha Letman, Akhil Mathur, Alan Schelten, Alex Vaughan, et~al.
\newblock The llama 3 herd of models.
\newblock {\em arXiv preprint arXiv:2407.21783}, 2024.

\bibitem{xi2025sparse_efficient_dit_video_sparseattn}
Haocheng Xi, Shuo Yang, Yilong Zhao, Chenfeng Xu, Muyang Li, Xiuyu Li, Yujun Lin, Han Cai, Jintao Zhang, Dacheng Li, et~al.
\newblock Sparse videogen: Accelerating video diffusion transformers with spatial-temporal sparsity.
\newblock {\em arXiv preprint arXiv:2502.01776}, 2025.

\bibitem{xia2025training}
Yifei Xia, Suhan Ling, Fangcheng Fu, Yujie Wang, Huixia Li, Xuefeng Xiao, and Bin Cui.
\newblock Training-free and adaptive sparse attention for efficient long video generation.
\newblock {\em arXiv preprint arXiv:2502.21079}, 2025.

\bibitem{guo2024blocksparse}
Junxian Guo, Haotian Tang, Shang Yang, Zhekai Zhang, Zhijian Liu, and Song Han.
\newblock {Block Sparse Attention}.
\newblock \url{https://github.com/mit-han-lab/Block-Sparse-Attention}, 2024.

\bibitem{li2023qdiffusion}
Xiuyu Li, Yijiang Liu, Long Lian, Huanrui Yang, Zhen Dong, Daniel Kang, Shanghang Zhang, and Kurt Keutzer.
\newblock Q-diffusion: Quantizing diffusion models.
\newblock In {\em Proceedings of the IEEE/CVF International Conference on Computer Vision (ICCV)}, 2023.

\bibitem{zhang2025sageattention}
Jintao Zhang, Jia wei, Pengle Zhang, Jun Zhu, and Jianfei Chen.
\newblock Sageattention: Accurate 8-bit attention for plug-and-play inference acceleration.
\newblock In {\em The Thirteenth International Conference on Learning Representations}, 2025.

\bibitem{zhang2024sageattention2}
Jintao Zhang, Haofeng Huang, Pengle Zhang, Jia Wei, Jun Zhu, and Jianfei Chen.
\newblock Sageattention2: Efficient attention with thorough outlier smoothing and per-thread int4 quantization.
\newblock In {\em International Conference on Machine Learning (ICML)}, 2025.

\bibitem{li2024svdquant}
Muyang Li*, Yujun Lin*, Zhekai Zhang*, Tianle Cai, Xiuyu Li, Junxian Guo, Enze Xie, Chenlin Meng, Jun-Yan Zhu, and Song Han.
\newblock Svdquant: Absorbing outliers by low-rank components for 4-bit diffusion models.
\newblock In {\em The Thirteenth International Conference on Learning Representations}, 2025.

\bibitem{xie2025sana}
Enze Xie, Junsong Chen, Junyu Chen, Han Cai, Haotian Tang, Yujun Lin, Zhekai Zhang, Muyang Li, Ligeng Zhu, Yao Lu, and Song Han.
\newblock {SANA}: Efficient high-resolution text-to-image synthesis with linear diffusion transformers.
\newblock In {\em The Thirteenth International Conference on Learning Representations}, 2025.

\bibitem{chen2025deep}
Junyu Chen, Han Cai, Junsong Chen, Enze Xie, Shang Yang, Haotian Tang, Muyang Li, and Song Han.
\newblock Deep compression autoencoder for efficient high-resolution diffusion models.
\newblock In {\em The Thirteenth International Conference on Learning Representations}, 2025.

\bibitem{li2023snapfusion}
Yanyu Li, Huan Wang, Qing Jin, Ju~Hu, Pavlo Chemerys, Yun Fu, Yanzhi Wang, Sergey Tulyakov, and Jian Ren.
\newblock Snapfusion: Text-to-image diffusion model on mobile devices within two seconds.
\newblock In {\em Thirty-seventh Conference on Neural Information Processing Systems}, 2023.

\bibitem{yin2024improved}
Tianwei Yin, Micha{\"e}l Gharbi, Taesung Park, Richard Zhang, Eli Shechtman, Fredo Durand, and William~T. Freeman.
\newblock Improved distribution matching distillation for fast image synthesis.
\newblock In {\em The Thirty-eighth Annual Conference on Neural Information Processing Systems}, 2024.

\bibitem{child2019generating}
Rewon Child, Scott Gray, Alec Radford, and Ilya Sutskever.
\newblock Generating long sequences with sparse transformers.
\newblock {\em arXiv preprint arXiv:1904.10509}, 2019.

\bibitem{xiao2025duoattention}
Guangxuan Xiao, Jiaming Tang, Jingwei Zuo, junxian guo, Shang Yang, Haotian Tang, Yao Fu, and Song Han.
\newblock Duoattention: Efficient long-context {LLM} inference with retrieval and streaming heads.
\newblock In {\em The Thirteenth International Conference on Learning Representations}, 2025.

\bibitem{xu2025xattention_efficient_sparseattn}
Ruyi Xu, Guangxuan Xiao, Haofeng Huang, Junxian Guo, and Song Han.
\newblock Xattention: Block sparse attention with antidiagonal scoring.
\newblock {\em arXiv preprint arXiv:2503.16428}, 2025.

\bibitem{li2023distrifusion}
Muyang Li, Tianle Cai, Jiaxin Cao, Qinsheng Zhang, Han Cai, Junjie Bai, Yangqing Jia, Ming-Yu Liu, Kai Li, and Song Han.
\newblock Distrifusion: Distributed parallel inference for high-resolution diffusion models.
\newblock In {\em Proceedings of the IEEE/CVF Conference on Computer Vision and Pattern Recognition (CVPR)}, 2024.

\bibitem{li2024svdqunat}
Muyang Li, Yujun Lin, Zhekai Zhang, Tianle Cai, Xiuyu Li, Junxian Guo, Enze Xie, Chenlin Meng, Jun-Yan Zhu, and Song Han.
\newblock Svdqunat: Absorbing outliers by low-rank components for 4-bit diffusion models.
\newblock {\em arXiv preprint arXiv:2411.05007}, 2024.

\bibitem{liu2024timestep}
Feng Liu, Shiwei Zhang, Xiaofeng Wang, Yujie Wei, Haonan Qiu, Yuzhong Zhao, Yingya Zhang, Qixiang Ye, and Fang Wan.
\newblock Timestep embedding tells: It's time to cache for video diffusion model, 2024.

\bibitem{huang2023vbench}
Ziqi Huang, Yinan He, Jiashuo Yu, Fan Zhang, Chenyang Si, Yuming Jiang, Yuanhan Zhang, Tianxing Wu, Qingyang Jin, Nattapol Chanpaisit, Yaohui Wang, Xinyuan Chen, Limin Wang, Dahua Lin, Yu~Qiao, and Ziwei Liu.
\newblock {VBench}: Comprehensive benchmark suite for video generative models.
\newblock In {\em Proceedings of the IEEE/CVF Conference on Computer Vision and Pattern Recognition}, 2024.

\bibitem{zhang2018perceptual}
Richard Zhang, Phillip Isola, Alexei~A Efros, Eli Shechtman, and Oliver Wang.
\newblock The unreasonable effectiveness of deep features as a perceptual metric.
\newblock In {\em CVPR}, 2018.

\end{thebibliography}

\clearpage
\newpage
\appendix

\section*{Appendix}

\section{Detailed Proof}\label{sec:app:proofs}

\subsection{Proof of Theorem~\ref{thm:draft_error}}

\begin{proof}
First, observe that for any $u \in R_i$ and $v \in R_j$, the draft attention assigns
\begin{align}
(S_{\mathrm{draft}})_{uv} = \widetilde S_{ij}, \quad \text{while} \quad S_{uv} = \langle Q_u, K_v \rangle.
\end{align}

By the definition of $\delta$, we have
\begin{align}
|S_{uv} - (S_{\mathrm{draft}})_{uv}| = |S_{uv} - \widetilde S_{ij}| \le \delta.
\end{align}

Then, summing over all $n^2$ token pairs gives
\begin{align}
\| S - S_{\mathrm{draft}} \|_F^2 = \sum_{u,v} |S_{uv} - (S_{\mathrm{draft}})_{uv}|^2 \le n^2 \delta^2.
\end{align}

Taking square roots on both sides yields the desired result:
\begin{align}
\| S - S_{\mathrm{draft}} \|_F \le \delta n.
\end{align}
This completes the proof.
\end{proof}

\subsection{Proof of Theorem~\ref{thm:mask_error}}

\begin{proof}
The mask $\widehat M$ zeros out exactly $(1 - r)n^2$ entries corresponding to dropped blocks. For each $(u, v)$ in a dropped block, we have:
\begin{align}
  |S_{uv}| \le |S_{uv} - \widetilde S_{ij}| + |\widetilde S_{ij}| \le \delta + t.
\end{align}
Summing squared errors over $(1 - r)n^2$ entries yields the bound:
\begin{align}
  \| S - S \odot \widehat M \|_F^2 \le (1 - r)n^2(\delta + t)^2 \quad \Rightarrow \quad \| S - S \odot \widehat M \|_F \le n (\delta + t) \sqrt{1 - r}.
\end{align}
We then finish the proof.
\end{proof}

\end{document}